\documentclass[11pt,letterpaper]{article}
\usepackage[T1]{fontenc}
\usepackage{lmodern}
\usepackage[margin=1in]{geometry}
\usepackage{amsmath,amssymb,booktabs,array,graphicx,microtype}
\usepackage{placeins}
\usepackage[round,authoryear]{natbib}
\usepackage[hidelinks]{hyperref}
\hypersetup{pdftitle={Predicting Lost Updates Is Not Enough to Choose a Rounding Policy},
            pdfauthor={Zekai Shang}}
\title{Predicting Lost Updates Is Not Enough\\to Choose a Rounding Policy}
\author{Zekai Shang\\University of Illinois Urbana-Champaign\\\texttt{zekais3@illinois.edu}}
\date{}

\newcommand{\rne}{\textsf{RNE}}

\newcommand{\indicator}{\mathbf{1}}
\begin{document}
\maketitle
\begin{abstract}
In low-precision training, a weight can round back to its previous value
after a small update. We test whether predicting how often this happens
helps choose a rounding rule. In Pythia-410M continuations from a common checkpoint,
a higher-precision pilot accurately predicted the fraction of matrix weights
left unchanged at each of the first 64 updates. Across three schedules with optimizer
state retained or reset, nearest rounding gave lower validation loss at 256
updates than stochastic rounding or retaining 32-bit weights. The same ordering
held at 512 updates with retained state. Accurately forecasting the unchanged fraction did not identify a
benefit from switching policies in these runs.
\end{abstract}

\section{Introduction}

A training step can leave a low-precision weight unchanged when the proposed
value rounds back to its stored value. Stochastic rounding gives a small
nonzero update a chance to move the weight to a neighboring value. A
higher-precision copy can accumulate small updates \citep{Micikevicius2018,Gupta2015,Zamirai2020,ECO2026}.
We ask whether the frequency of unchanged weights helps determine when either
alternative improves training.

A short higher-precision reference run could help select a rounding policy
for several later runs. We project its weights onto the low-precision grid
and count proposals that round back to their starting values. Running each
candidate policy briefly and comparing validation loss provides a direct
alternative. When low precision is required only for deployment, the finished
model can be quantized.

We compare nearest-even rounding (RNE), stochastic rounding (SR) and unrestricted
FP32 weight updates in continuations from one Pythia-410M checkpoint. The three
policies share initial weights and minibatches within each schedule. A reference
run on separate minibatches accurately predicts that most weights will remain
unchanged. RNE gives the lowest validation loss on all three schedules with Adam
state either retained or reset. In our earlier GPT-2-family experiments from
scratch, SR gave lower loss than RNE.

\section{Experimental design}
\label{sec:methods}

\subsection{Rounding policies and optimizer state}

We adapt the released Pythia-410M-deduped step-129000 checkpoint
\citep{Biderman2023Pythia} for 512 FP32 weight updates, reducing development
loss from 2.518533 to 2.429709. Before running the continuations, we select the
first three of six prepared source--target schedule pairs for the compute
budget and specify both optimizer conditions. Each pair defines a block with
two 64-update reference runs and six 256-update target runs, for 24 runs in total.

All runs start from the same E4M3FN-projected checkpoint. To test sensitivity to
optimizer state, we either retain Adam's moments and update counter 512 or reset
both to zero. Both conditions continue the same learning-rate schedule from
step 129512. Target policies share minibatches within each block;
the separately seeded reference schedule is also shared across optimizer
conditions. We exclude training windows that overlap other schedules,
adaptation or development data.

After each update, the policies retain the updated FP32 matrix weights or round
them using RNE or SR. Rounding applies to the same 405,012,480 matrix coordinates;
one-dimensional parameters remain unrestricted. We store parameters and AdamW
moments in FP32 and use BF16 parameters and buffers for computation with eager
attention. SR uses one private random stream per block, paired across optimizer
conditions. Appendix~\ref{app:protocol} gives the quantizer and full settings.

We evaluate all targets on the same 128 contexts, excluded from adaptation and
runtime calibration. Initial context losses match exactly, with mean 2.433685
nats per token. The pre-specified primary analysis measures how reset changes
the RNE--FP32 loss difference:
\begin{equation}
 I_{\rne,b}=\bigl(L_{\rne,b}-L_{\mathrm{FP32},b}\bigr)_{\rm reset}
             -\bigl(L_{\rne,b}-L_{\mathrm{FP32},b}\bigr)_{\rm preserved},
 \label{eq:reset-interaction}
\end{equation}
where $L$ is final validation cross entropy. Positive $I_{\rne,b}$ means reset
increases RNE loss relative to the FP32 baseline. The corresponding SR
interaction is secondary.

After observing the 256-update results, we repeat each preserved-state
comparison for 512 updates, retaining the original SR stream and adding a
second. These twelve runs reuse the first 256 minibatches and 128 evaluation
contexts; appended minibatches and new rounding seeds were specified before
the extension runs
(Appendix~\ref{app:horizon}).

\subsection{What the reference predicts}

Let $Q$ be the target nearest-even quantizer, $q_i$ a stored grid value, and
$d_i$ the optimizer's full signed proposal, including decay. An unchanged
coordinate satisfies
\begin{equation}
 Q(q_i+d_i)=q_i.
 \label{eq:event}
\end{equation}
For a higher-precision source trajectory $(w^r_t,d^r_t)$, project
$\widehat q_{t,i}=Q(w^r_{t,i})$ and predict
\begin{equation}
 X_t=\frac1N\sum_{i=1}^{N}
 \indicator\!\left[Q(\widehat q_{t,i}+d^r_{t,i})=\widehat q_{t,i}\right].
 \label{eq:proxy}
\end{equation}
The observed target fraction $Y_t$ counts weights unchanged by actual
write-back. Both fractions include zero proposals and are computed per update.

We evaluate the reference over its 64 updates and the first 64 RNE target
updates. The pre-specified secondary metrics are raw and time-centered RMSE
against $Y_t$, compared with predicting every coordinate unchanged.
Centering subtracts each trajectory's mean but preserves shared temporal shape.

\section{Validation loss under each rounding policy}
\label{sec:policy-results}

\begin{figure}[!ht]
\centering
\includegraphics[width=\linewidth]{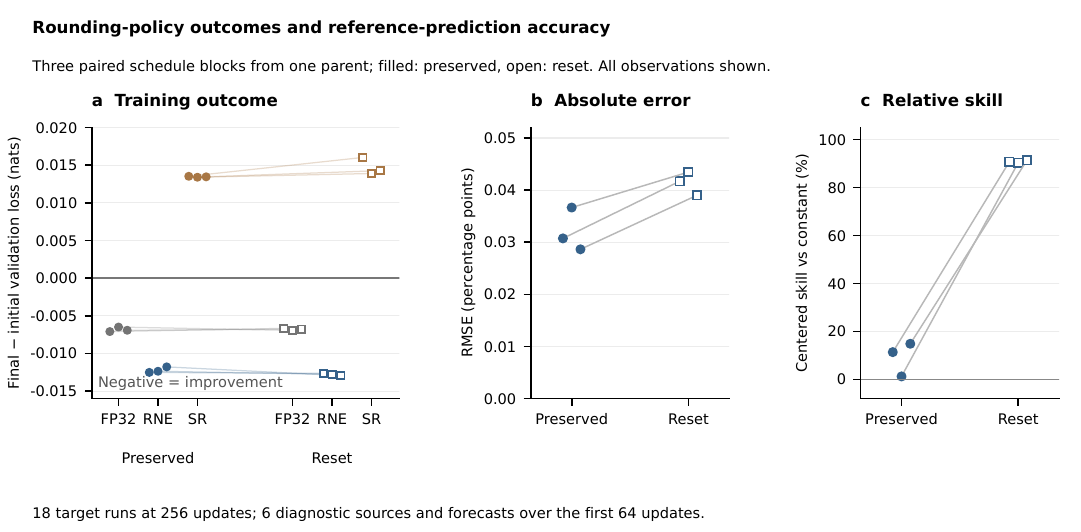}
\caption{Policy outcomes and prospective forecast diagnostics.
\textbf{a,} All eighteen 256-update loss changes; negative means improvement.
\textbf{b,} Raw 64-update reference RMSE by block and optimizer condition.
\textbf{c,} Centered RMSE skill $1-E_c/\operatorname{SD}_t(Y)$
(Eq.~\eqref{eq:relative-skill}). Markers show individual runs or paired
conditions.}
\label{fig:policy-and-prediction}
\end{figure}

Table~\ref{tab:policy-endpoints} reports final validation loss for all eighteen
targets after 256 updates. RNE has the lowest loss in every block and optimizer
condition (Fig.~\ref{fig:policy-and-prediction}a). FP32 improves from 2.433685
to 2.426590--2.427189, and RNE improves to 2.420748--2.421899.
SR finishes at 2.447088--2.449715, above initialization. Relative to matched FP32,
RNE is lower by 0.004863--0.006171 nats. Relative to RNE, SR is higher by
0.025241--0.028678 nats.

\begin{table}[!ht]
\caption{All eighteen target endpoints after 256 updates from initial loss
2.433685. Lower is better.}
\label{tab:policy-endpoints}
\centering\small
\begin{tabular}{clrrr}
\toprule
Block & Optimizer & FP32 & RNE & SR\\
\midrule
1 & Preserved & 2.426590 & 2.421168 & 2.447213\\
1 & Reset & 2.427018 & 2.421036 & 2.449715\\
2 & Preserved & 2.427189 & 2.421319 & 2.447088\\
2 & Reset & 2.426744 & 2.420908 & 2.447584\\
3 & Preserved & 2.426762 & 2.421899 & 2.447140\\
3 & Reset & 2.426919 & 2.420748 & 2.447963\\
\bottomrule
\end{tabular}
\end{table}

Reset changes the RNE--FP32 loss difference (Eq.~\eqref{eq:reset-interaction}) by
$-0.000560$, $+0.000034$ and $-0.001308$ nats in blocks 1--3: signs are mixed,
and the policy ordering is unchanged. The corresponding SR interactions are
$+0.002073$, $+0.000940$ and $+0.000667$ nats.

At 512 updates, RNE still has the lowest loss in all three preserved-state
blocks (Appendix~\ref{app:horizon}). Its loss is lower than FP32 by
0.002797--0.004940 nats; both improve from initialization. SR is higher than
RNE by 0.035621--0.038231 nats and worsens from initialization.

In a retrospective comparison, RNE had the lowest loss both at the first
scheduled evaluation (update 64) and at update 256 across all three schedules
and both optimizer conditions. Both evaluations used the same 128 contexts.

\section{Forecast accuracy and shared temporal structure}
\label{sec:forecast-results}

\begin{figure}[!t]
\centering
\includegraphics[width=\linewidth]{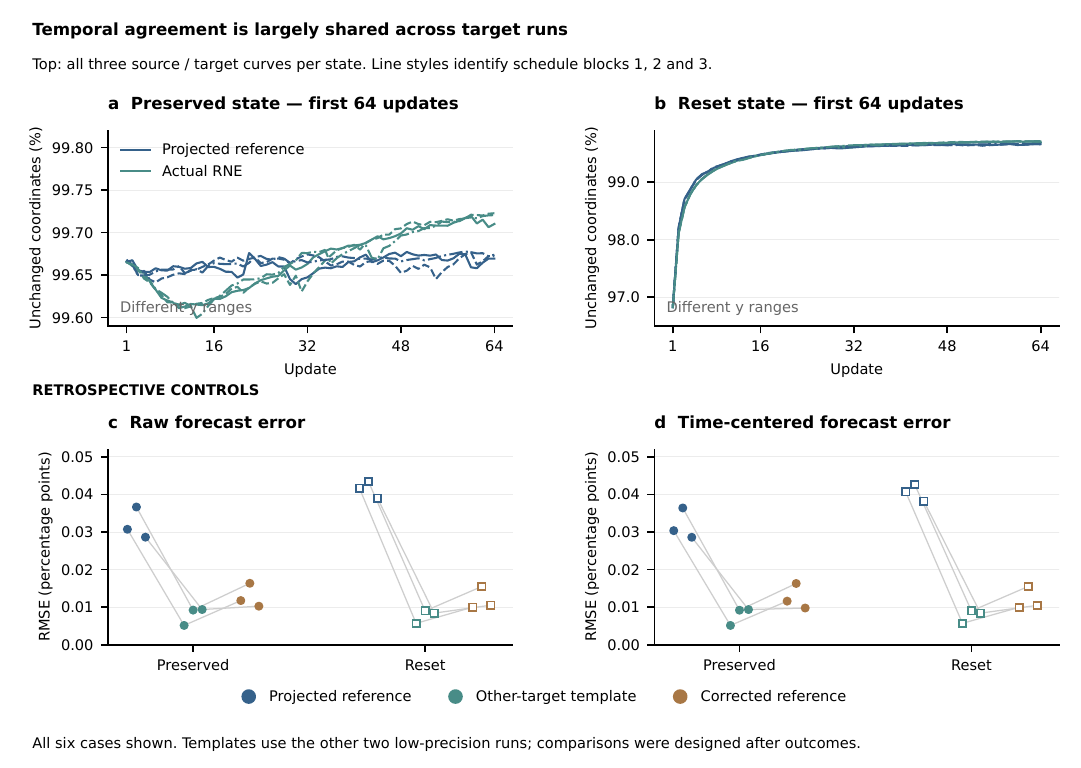}
\caption{Retrospective temporal controls. Upper panels show all source and
target trajectories, with separate scales by optimizer condition. Lower panels
compare raw and centered errors for every block. Target and source templates
average the other two RNE or higher-precision trajectories, respectively,
excluding the evaluated block.}
\label{fig:temporal-controls}
\end{figure}

\subsection{Relative skill and absolute error diverge}

For the fraction of unchanged weights, forecast RMSE is 0.000286--0.000367 with preserved state and
0.000390--0.000435 after reset, below the always-unchanged baseline in every case
(Fig.~\ref{fig:policy-and-prediction}b). Reset worsens absolute error in each
block: mean raw RMSE rises from 0.0003201 to 0.0004139 and centered RMSE from
0.0003179 to 0.0004052.

Normalized temporal skill moves in the opposite direction. Define
\begin{equation}
 E_c=\sqrt{\frac1H\sum_{t=1}^{H}
       \bigl[(X_t-\bar X)-(Y_t-\bar Y)\bigr]^2},\qquad
 S_c=1-\frac{E_c}{\operatorname{SD}_t(Y)},\quad H=64.
 \label{eq:relative-skill}
\end{equation}
Centered RMSE skill ranges from 0.013--0.149 with preserved state to 0.904--0.914
after reset. Mean target temporal standard deviation rises from 0.000349 to
0.004427, about $12.68\times$, increasing relative skill despite greater
absolute error.

With preserved state, the mean unchanged fraction moves from 0.996655 to
0.997178. After reset it rises from 0.968235 to 0.997108. The reference follows
this transient, making the constant baseline less accurate.
Relative skill alone therefore does not establish target-specific information.

\subsection{Retrospective diagnosis with other low-precision runs}
\label{sec:retrospective}

In a retrospective analysis, we predict each block by averaging the other two
completed RNE target trajectories under the same optimizer condition. This
leave-one-block-out template measures how much of the trajectory is shared.

The template has lower raw and centered RMSE than the reference in all six
comparisons (Fig.~\ref{fig:temporal-controls}). Its mean raw RMSE across blocks
is $7.94\times10^{-5}$ with preserved state and $7.73\times10^{-5}$ after reset,
versus reference errors $3.20\times10^{-4}$ and $4.14\times10^{-4}$.
The reference's raw RMSE is 3.05--7.28 times the template's.
Adding the difference between the block's reference and the mean of the other
two references to the target template increases error in every case
(Appendix~\ref{app:temporal-controls}).
A descriptive decomposition attributes 97.57\% of preserved-state and 99.99\%
of reset-state temporal variation to a common time effect. Removing the first
32 updates lowers reset mean centered skill from 0.908 to 0.397; the preserved
mean is 0.043. This exploratory analysis retains all blocks and shows
how much the full-window score depends on the initial transient
(Appendix~\ref{app:temporal-controls}).

\section{Earlier reference forecasts and interventions}
\label{sec:original}

The earlier Pythia audit tests whether the reference improves prediction over
simple baselines. Fifteen holdouts
at three checkpoint ages share three fresh sources, one per age, and three
historical reference/target pairs per age. A geometric-margin selector meets absolute
error/coverage requirements on every holdout but has higher error than predicting
unchanged on the same accepted events. A unit-gain source correction has higher raw and centered RMSE
than the historical template in all 60 dependent target/template comparisons.
Both original and amended joint decisions are negative.
Appendix~\ref{app:original-audit} details selection
limits and the post-outcome code repair.

The new reset comparison uses a common projected start and a different
runtime. How much restarting contributed to the earlier forecast error remains
open. In the earlier runs, loss rose for eight early/middle FP32 references
and fell for all 33 RNE targets, whose initializations differed from those of
the references (Appendix~\ref{app:supporting}).

In a secondary comparison on three NeoX targets \citep[Section~4.3]{Shang2026v4}, historical low-precision curves
predicted the fraction of unchanged weights with mean RMSE 0.00360. Correcting
that forecast with one shared reference raised the error to 0.00438. Using a
reference matched to each target lowered it to 0.00283, at the cost of one
high-precision run per target.

We also tested whether a pilot could choose which 10\% of weight tiles to
keep in higher precision. On one exploratory schedule, selecting tiles by the
pilot's estimated rounding damage gave loss 2.80131 in 272 seconds including
the pilot. Selecting the same number at random gave 2.80100 in 205 seconds.
Both improved on RNE's 2.80954 (Table~\ref{tab:development-all}).
When training GPT-2-family models from scratch for 10,000 updates, SR recovered
roughly 89--92\% of the observed RNE--FP32 loss gap (Appendix~\ref{app:original-audit}).

\section{Related work}

Master-weight accumulation, stochastic rounding and compensated updates
address lost low-precision updates
\citep{Micikevicius2018,Gupta2015,Zamirai2020,Connolly2021SR,Ozkara2025,ECO2026}.
Other methods use online measurements of update visibility
\citep{Xia2024,Yu2026,PULSE2026,MAdam2026}. We evaluate whether predicting these
events improves the choice of rounding policy.

Selective classification studies coverage/error tradeoffs
\citep{ElYaniv2010Selective}. Our event-level comparison measures error on the
same accepted examples for both the reference and the majority label. Prior
work also uses saliency to guide precision allocation
\citep{Molchanov2017,Dong2019HAWQ,Fu2020FracTrain}.

\section{Conclusion}

RNE achieves lower validation loss than FP32 and SR in these matched
continuations, despite accurate forecasts that most weights remain unchanged.
Reset increases relative forecast skill while worsening absolute error, and
other-run templates have lower prediction error. Lost-update counts alone
therefore do not establish training harm; the relevant test is whether changing
the policy improves training.

\section*{Reproducibility statement}
The ancillary files contain per-update trajectories, validation losses and
Python scripts to reproduce the reported comparisons for all 36 Pythia-410M
reference and target runs. The appendix gives the experimental settings
and analysis timing.

\section*{AI use statement}
Generative AI assisted with hypothesis refinement, mathematical and claim
review, experimental design, implementation, interpretation, literature search,
drafting, editing, figures and consistency checks. Reported quantities were
checked against retained artifacts, and AI-assisted code was tested.

\FloatBarrier
\bibliography{references}
\bibliographystyle{plainnat}
\appendix
\section{Continuation experiment: setup and complete results}
\label{app:protocol}

We train for 512 updates with unrestricted FP32 weights, starting
from released Pythia-410M-deduped step-129000 weights. The 64-context
development loss decreases from 2.518533 to 2.429709, meeting the predefined
checkpoint-selection criterion. We evaluate continuations on 128 held-out
contexts excluded from adaptation and runtime calibration, with 2048
prediction tokens each, or 262,144 tokens. All targets and sources start from
the same projected adapted checkpoint and have identical initial losses on
each context.

We use AdamW with betas $(0.9,0.95)$, epsilon $10^{-8}$, matrix weight decay
0.1 and global gradient clipping at 1.0. Each update processes 66 length-2048
contexts in microbatches of one. Peak/minimum learning rates are
$3\times10^{-4}/3\times10^{-5}$, with warmup 1430 and cosine horizon 143000
at virtual step $129512+u$. In reset mode, we zero both moments and the Adam
counter while retaining this schedule. Preserved mode retains moments learned
before the common projection.

We use an analytic E4M3FN quantizer with scale 128, nearest-even ties,
saturation at physical values $\pm3.5$, subnormal spacing and canonical
numerical zero. Quantized policies round all 405,012,480 trainable 2-D
coordinates after each update and leave other parameters unrestricted.
Software SR rounds between adjacent finite codes, with conditional mean equal
to the clipped proposal.
Each block uses one private SR stream shared across optimizer modes.

We store parameters, gradients and optimizer state in FP32. For BF16 eager
forward/backward computation, we temporarily cast parameter values, round
registered floating buffers once and retain them in BF16, and accumulate
gradients in FP32. We use PyTorch 2.8.0+cu128 and Transformers 4.56.2 on an
RTX 5090 and check the arithmetic exactly on small models. A 256-update target
with diagnostic logging takes 2881--2900 seconds. Acquiring a 64-update
reference adds 746--750 seconds to the cost of evaluating the forecast.

We selected three schedule pairs to fit the compute budget before observing
continuation results. Within each block, all target policies and optimizer
modes use the same minibatches; sources use a separate schedule shared across
optimizer modes. We sample training windows with recorded seeds and reject
overlap with other source/target, adaptation and development windows. All 24
runs completed: six 64-update sources and eighteen 256-update targets.
We specified the primary reset interaction in Eq.~\eqref{eq:reset-interaction},
secondary SR interaction and reference-forecast metrics before these runs.
The three schedule blocks share one adapted checkpoint, with paired optimizer
modes.

\begin{table}[!ht]
\caption{All eighteen target outcomes after 256 updates from initial loss
2.433685. Negative change is improvement.}
\label{tab:all-target-outcomes}
\centering\small
\begin{tabular}{cllrr}
\toprule
Block & Optimizer & Policy & Final loss & Change\\
\midrule
1 & Preserved & FP32 & 2.426590 & $-0.007095$\\
1 & Preserved & RNE & 2.421168 & $-0.012517$\\
1 & Preserved & SR & 2.447213 & $+0.013528$\\
1 & Reset & FP32 & 2.427018 & $-0.006667$\\
1 & Reset & RNE & 2.421036 & $-0.012649$\\
1 & Reset & SR & 2.449715 & $+0.016029$\\
2 & Preserved & FP32 & 2.427189 & $-0.006496$\\
2 & Preserved & RNE & 2.421319 & $-0.012366$\\
2 & Preserved & SR & 2.447088 & $+0.013403$\\
2 & Reset & FP32 & 2.426744 & $-0.006941$\\
2 & Reset & RNE & 2.420908 & $-0.012777$\\
2 & Reset & SR & 2.447584 & $+0.013899$\\
3 & Preserved & FP32 & 2.426762 & $-0.006923$\\
3 & Preserved & RNE & 2.421899 & $-0.011786$\\
3 & Preserved & SR & 2.447140 & $+0.013455$\\
3 & Reset & FP32 & 2.426919 & $-0.006766$\\
3 & Reset & RNE & 2.420748 & $-0.012937$\\
3 & Reset & SR & 2.447963 & $+0.014278$\\
\bottomrule
\end{tabular}
\end{table}

\begin{table}[!ht]
\caption{Effect of optimizer reset on each policy's loss relative to FP32.}
\label{tab:all-reset-interactions}
\centering\small
\begin{tabular}{crr}
\toprule
Block & Reset effect on RNE--FP32 & Reset effect on SR--FP32\\
\midrule
1 & $-0.000560$ & $+0.002073$\\
2 & $+0.000034$ & $+0.000940$\\
3 & $-0.001308$ & $+0.000667$\\
\bottomrule
\end{tabular}
\end{table}

\begin{table}[!ht]
\caption{Source-forecast diagnostics over 64 updates. $S_c$ is centered
RMSE skill (Eq.~\eqref{eq:relative-skill}).}
\label{tab:all-forecast-diagnostics}
\centering\small
\begin{tabular}{clrrrr}
\toprule
Block & Optimizer & RMSE & Centered RMSE & SD$(Y)$ & $S_c$\\
\midrule
1 & Preserved & 0.0003073 & 0.0003037 & 0.0003427 & 0.1138\\
1 & Reset & 0.0004169 & 0.0004067 & 0.0044251 & 0.9081\\
2 & Preserved & 0.0003666 & 0.0003640 & 0.0003687 & 0.0125\\
2 & Reset & 0.0004346 & 0.0004264 & 0.0044215 & 0.9036\\
3 & Preserved & 0.0002864 & 0.0002861 & 0.0003360 & 0.1487\\
3 & Reset & 0.0003902 & 0.0003824 & 0.0044349 & 0.9138\\
\bottomrule
\end{tabular}
\end{table}

\section{Retrospective temporal controls}
\label{app:temporal-controls}

We constructed these controls after observing the results.
Let $X_b$ be block $b$'s projected-reference curve and $Y_b$ its RNE target
curve, both over the first 64 updates. Under the same optimizer condition,
define
\begin{equation}
 B_{Y,-b}=\frac12\sum_{j\ne b}Y_j,\qquad
 B_{X,-b}=\frac12\sum_{j\ne b}X_j,\qquad
 H_b=B_{Y,-b}+X_b-B_{X,-b}.
 \label{eq:lobo-controls}
\end{equation}
Both templates exclude the evaluated block. The target template uses other
low-precision runs, including those completed later.

\begin{table}[!ht]
\caption{Mean raw RMSE across three blocks by optimizer condition.
Individual comparisons are retained in source data.}
\label{tab:temporal-controls-summary}
\centering\small
\begin{tabular}{lrr}
\toprule
Predictor & Preserved & Reset\\
\midrule
Projected reference $X_b$ & 0.000320100 & 0.000413887\\
Other-run target template $B_{Y,-b}$ & 0.000079376 & 0.000077279\\
Unit-gain source hybrid $H_b$ & 0.000128156 & 0.000120421\\
Always unchanged & 0.003345823 & 0.006820491\\
\bottomrule
\end{tabular}
\end{table}

The target template has lower raw and centered RMSE than the reference and
unit-gain hybrid in every comparison. Reference/template RMSE ratios are
3.048--7.280 raw and 3.048--7.102 centered.

To measure shared temporal structure, center each target over time and write
$y_{b,t}=Y_{b,t}-\overline Y_b$. Let $\bar y_t$ average the three centered
curves. The common-time fraction is
\begin{equation}
 \frac{3\sum_t\bar y_t^2}{\sum_b\sum_t y_{b,t}^2}.
\end{equation}
It is 0.975749 with preserved state and 0.999860 after reset.
In an exploratory sensitivity analysis, we remove 0, 1, 4,
8, 16 or 32 initial updates from all blocks. Removing 32 gives reset
centered skill 0.297--0.455 (mean 0.397) and preserved skill
$-0.103$--0.130 (mean 0.043). We chose these windows after observing the results.

\section{Original forecast evaluation and rounding experiments}
\label{app:original-audit}

The original Pythia study evaluates 45 runs of 512 updates at released steps
7000, 72000 and 129000. Each age has three historical reference/target pairs,
one additional source, three development targets and five holdouts. References
start from unquantized weights and targets from grid-projected weights;
both reset Adam moments and counters while retaining checkpoint-age learning
rates. Fifteen holdouts share three sources, and the 60 full/leave-one-out
template comparisons are dependent.
Original targets store decoded E4M3 values in FP32 with BF16
computation under FSDP/PyTorch 2.12.1/Transformers 5.12.1, unlike the new
single-GPU runtime.

The selector predicts projected code equality using normalized distance to
the destination-cell boundary. We select the lowest of 17 thresholds meeting
all nine development targets' error/coverage limits before evaluating holdouts.
Bin 32 requires a geometric margin of at least $1/8$. We estimate event risk
from $2^{20}$ sampled coordinate trajectories, held fixed over 512 updates.
The temporal forecast pools all 405,012,480 eligible coordinates. Every holdout
meets the error upper bound of 0.059 and coverage lower bound of 0.50. Always
predicting unchanged on the selector's accepted events gives lower empirical
error on every holdout. Only 14.2--22.7\% of accepted change predictions are
correct.

The original bounds do not adjust for selecting among 17 thresholds on the
same coordinate sample, so the reported passes lack a selection-valid
simultaneous 95\% guarantee. The prospectively specified amendment adjusts
over the grid within each target.

The temporal correction combines a new source with historical source and
target means. It has higher raw and centered RMSE than history in every
comparison, with raw ratios of 1.703--5.475. Both the original evaluation and
the amendment adding event-baseline skill and material-effect requirements
return \texttt{NO\_GO}. After observing the results, we corrected the program's
links between holdouts and development runs before applying the existing
checks of input identities. Inputs, statistical formulas, thresholds and decisions
were unchanged; the reported results use the corrected executable.

Fitting a scalar oracle separately to each observed target for
$B_Y+\alpha(X_S-B_X)$ gives
maximal RMSE reductions of 3.3785\% raw and 3.3070\% centered, below the fixed
10\% requirement in every comparison for that correction direction.

In the 72-cell controlled timing grid, 55 cells have measured and predicted
post-initialization crossings; 52 of these fall within a descriptive 15\% band.
Four cells disagree on crossing category. A separate functional test finds models
majority-unchanged from the first update that still achieve 52.4--59.6\% of
reference progress.

When training GPT-2-family models from scratch for 10,000 updates, SR
recovers 89.24\% and 89.43\% of the RNE--FP32 loss gap across two seeds
with E5M2, and 92.16\% and 92.07\% with E4M3. RNE and SR share projected
initial codes. The FP32 baseline used to normalize the loss gap starts from
unquantized weights. Version~4 of this preprint reports the earlier native low-precision and
allocation experiments, including Study~1, which was inconclusive under its
registered criteria \citep[Sections~4--5]{Shang2026v4}.

\section{Reference quality and exploratory precision allocation}
\label{app:supporting}

We retrospectively analyze logged validation losses for all twelve unrestricted
references and thirty-three RNE targets on the same 135168-token validation
set. Loss increases for all eight early/middle references and decreases for
all four late references and thirty-three targets. Nine historical pairs share
training schedules, but references start from unquantized weights and targets
from projected weights, preventing attribution to rounding alone.

\begin{figure}[!ht]
\centering
\includegraphics[width=\linewidth]{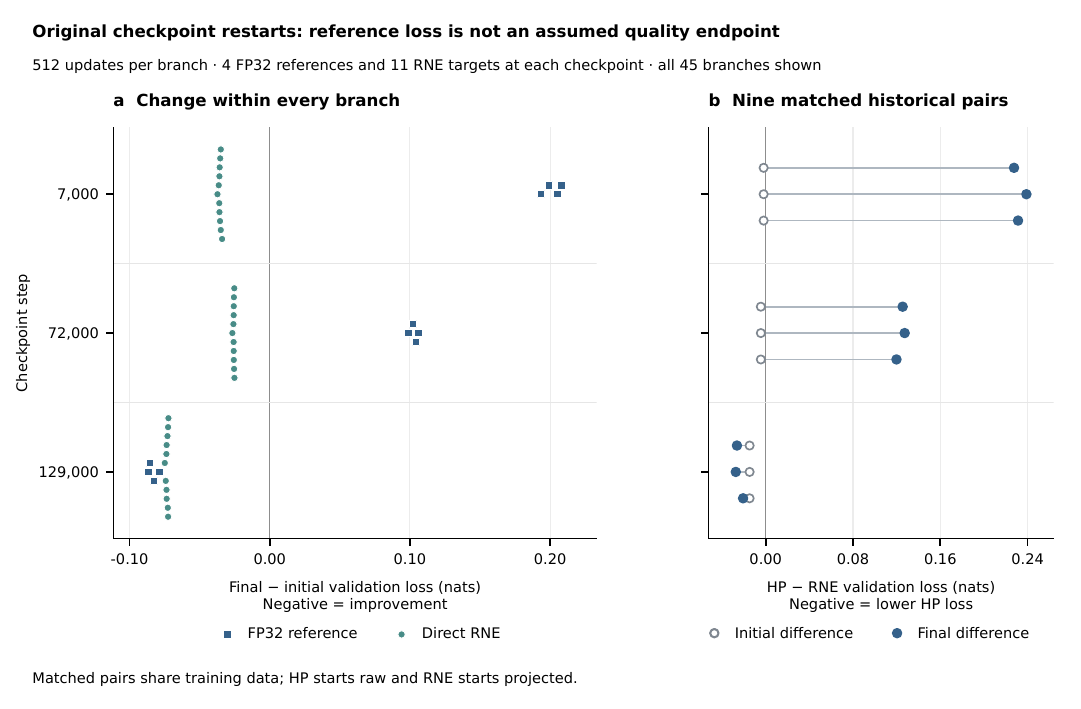}
\caption{Retrospective loss audit: all 45 branch changes and nine historical
pairs. Negative change means improvement. Reference/target initializations
differ through grid projection.}
\label{fig:original-health-support}
\end{figure}

The exploratory allocation study uses one early checkpoint with reset state,
eight length-512 contexts per batch, a 64-update pilot and 256 target updates.
All thirteen policies share one target schedule and 128 validation contexts.
HP and LP source pilots share batches distinct from the target's. The budget
selects 39552 of 395520 complete 1024-coordinate tiles. We store matrices in
FP8 with temporary BF16 linear computation and FP32 gradients/moments, unlike
the new policy study's FP32 storage. The signed rounding-damage score is a
first-order surrogate; its selected mask overlaps ordinary absolute proposal
saliency by 97.26\%.

\begin{table}[!ht]
\caption{All thirteen exploratory allocation policies on one target schedule.
Time includes required acquisition; peak memory is the maximum across source
and target.}
\label{tab:development-all}
\centering\small
\begin{tabular}{lrrr}
\toprule
Policy & Final loss & Time (s) & Peak (GiB)\\
\midrule
HP source, damage score & 2.801313 & 272.0 & 8.467\\
LP source, damage score & 2.801303 & 275.8 & 7.740\\
HP source, absolute saliency & 2.801371 & 262.4 & 8.467\\
HP source, freeze frequency & 2.809637 & 255.0 & 8.467\\
Target LP pilot + restart & 2.801358 & 274.0 & 7.740\\
Online selection & 2.801688 & 211.3 & 7.740\\
Global random master mask & 2.800996 & 204.8 & 7.127\\
RNE & 2.809539 & 213.8 & 6.964\\
Full SR & 2.856198 & 213.9 & 7.540\\
Blanket masters & 2.827551 & 203.7 & 8.472\\
Vanilla error feedback & 2.827225 & 203.1 & 8.859\\
FP32, projected start & 2.844808 & 157.1 & 7.522\\
FP32, raw start & 2.842769 & 156.4 & 7.501\\
\bottomrule
\end{tabular}
\end{table}

The high- and low-precision damage scores both select masks that improve on
RNE. Random selection at the same budget gives a slightly lower loss. For
selective SR, we held each tensor's source-derived quota fixed and randomized
the choice of tiles within tensors. Using the low-precision score gave loss
2.807470, compared with 2.807268 for random selection. Both used 7.740 GiB
of peak memory.

\section{Runtime comparison and checkpoint selection}
\label{app:runtime}

The exploratory allocation experiments use FP32-autocast/SDPA, unlike the
original FSDP BF16/eager path. On identical released late weights and the
original 66 evaluation contexts, loss is 2.387227 with FP32-autocast SDPA, 2.488028 with
FP32-autocast eager, 2.489406 with BF16 buffers/eager and 2.489315 with BF16
parameters/buffers/eager. The archived initial loss is 2.489754.

A planned 90-run allocation replication was cancelled before evaluating
new blocks because its SDPA checkpoint's 512-update development loss increased
from 2.411850 to 2.422852. A separately specified BF16/eager comparison
used the same released weights, batches, seed, optimizer and learning-rate
schedule; its development loss decreased from 2.518533 to 2.429709. We then
evaluated both adapted checkpoints under both runtimes on the original and
development contexts (all eight means below). The 128-context policy-study
evaluation set was excluded from checkpoint selection and runtime comparisons.

\begin{table}[!ht]
\caption{Validation loss for both adapted checkpoints under both runtimes.
Checkpoint ordering reverses with the evaluator.}
\label{tab:crossed-parents}
\centering\small
\begin{tabular}{llrr}
\toprule
Training path & Evaluation path & Original 66 & Health 64\\
\midrule
SDPA/autocast & SDPA/autocast & 2.397304 & 2.422852\\
SDPA/autocast & BF16/eager & 2.411034 & 2.437105\\
BF16/eager & SDPA/autocast & 2.405582 & 2.427929\\
BF16/eager & BF16/eager & 2.407106 & 2.429709\\
\bottomrule
\end{tabular}
\end{table}

We chose BF16/eager evaluation before running the policy comparison and use
it for all policies.

\FloatBarrier
\section{Longer continuations with a second stochastic-rounding stream}
\label{app:horizon}

We tested whether doubling the training horizon or changing the SR seed
altered the ordering of the three weight-update policies. After observing the
256-update results, we fixed the schedules, SR seeds and 512-update endpoint
for twelve runs with preserved Adam state. Within each block, FP32, RNE
and both SR runs used the same projected checkpoint, Adam moments and counter
512, learning rates, training seed and BF16/eager computation
(Appendix~\ref{app:protocol}). All four runs used the original 256 minibatches
followed by 256 new minibatches. The new training windows exclude earlier
windows and windows in the other blocks. We used the same 128 validation
contexts as at 256 updates.

In blocks 1--3, SR0 used the original rounding seeds 112247105, 112247205
and 112247305. SR1 used 896371238, 4086793475 and 2290992039.

\begin{table}[!ht]
\caption{All twelve preserved-state endpoints at update 512, from initial
loss 2.433685. SR0 retains the original rounding stream; SR1 uses the second
fixed stream. Lower is better.}
\label{tab:horizon-endpoints}
\centering\small
\begin{tabular}{crrrr}
\toprule
Block & FP32 & RNE & SR0 & SR1\\
\midrule
1 & 2.425002 & 2.420062 & 2.457502 & 2.457969\\
2 & 2.422727 & 2.419930 & 2.457391 & 2.455551\\
3 & 2.424446 & 2.420122 & 2.458353 & 2.456104\\
\bottomrule
\end{tabular}
\end{table}

RNE gave the lowest validation loss after 512 updates in each block
(Table~\ref{tab:horizon-endpoints}). Relative to FP32, RNE lowered loss by
0.004940, 0.002797 and 0.004324 nats per token in blocks 1--3. Relative to
RNE, SR0 raised loss by 0.037440, 0.037461 and 0.038231 nats, and SR1 by
0.037906, 0.035621 and 0.035983. The two SR streams exchanged ranks in
block 1 between updates 256 and 512.

All twelve runs started with identical per-context losses and finished at
Adam step 1024. The nine FP32, RNE and SR0 runs exactly reproduced the
earlier per-context losses at updates 0, 64, 128, 192 and 256 and the
fraction of unchanged weights at each of the first 256 updates.

\end{document}